\documentclass[conference]{IEEEtran}
\usepackage[hyphens]{url}
\usepackage{hyperref}
\hypersetup{breaklinks=true,colorlinks,allcolors=blue}
\usepackage{doi}
\usepackage{booktabs}
\usepackage{cite}
\usepackage{caption}
\usepackage{algorithmic}
\usepackage{graphicx}
\usepackage{textcomp}
\usepackage{xcolor}
\usepackage{amsmath}
\usepackage{balance}
\usepackage[left=1.62cm,right=1.62cm,top=1.9cm]{geometry}
\usepackage{graphicx}
\usepackage{subcaption}
\usepackage{makecell}

\def\BibTeX{{\rm B\kern-.05em{\sc i\kern-.025em b}\kern-.08em
    T\kern-.1667em\lower.7ex\hbox{E}\kern-.125emX}}
    
\begin{document}

\newcommand{\ahmed}[1]{\textcolor{magenta}{Dr. Ahmed: {#1}}}

\newcommand{\dana}[1]{\textcolor{red}{Dana: {#1}}}

\title{Hybrid Ensemble Learning for EEG-Based Epileptic Seizure Forecasting}

\author{\IEEEauthorblockN{Mason Dana and Khandaker Mamun Ahmed$^*$}
 \IEEEauthorblockA{\textit{The Beacom College of Computer and Cyber Sciences, Dakota State University} \\
Madison, SD, USA \\ $^*$Corresponding Author: khandakermamun.ahmed@dsu.edu}
}

\maketitle

\begin{abstract} 

Epileptic seizure forecasting aims to provide actionable warnings before seizure onset, yet patient-independent generalization and false-alarm control remain major challenges. We propose a calibrated hybrid ensemble for EEG-based seizure forecasting that combines five deep learning models and three classical machine learning models through a logistic regression stacking meta-learner. The proposed pipeline integrates signal preprocessing, handcrafted feature extraction, class-imbalance handling, probability calibration, and clinically motivated post-processing. We evaluate the framework on CHB-MIT using strict Leave-One-Patient-Out (LOPO) cross-validation, with threshold and post-processing parameters selected only on held-out meta data. On the filtered cohort, excluding patients with anomalous preictal rates below 1\% or above 15\%, the model achieves 74.2\% seizure-level sensitivity at 1.24 false alarms per hour, with an average warning time of 16.9 minutes. A test-tuned oracle constrained to the target false-alarm budget achieves 60.9\% sensitivity at 0.951 false alarms per hour, highlighting the importance of reporting sensitivity together with realized false-alarm rates. Our code is available at: \url{https://github.com/DanaMason/IEEE-CARS-Hybrid-Ensemble-Learning-for-EEG-Based-Epileptic-Seizure-Forecasting}


\end{abstract}


\section{Introduction}

Epilepsy affects more than 50 million individuals worldwide across the lifespan~\cite{frontiers_definitions_2024}. One of its most debilitating manifestations is the occurrence of unpredictable seizures, which arise from abnormal electrical activity in the brain. These events can substantially impair quality of life and are associated with driving restrictions, psychological distress, injury risk, and sudden unexpected death in epilepsy (SUDEP)~\cite{devinsky2016sudep}. Moreover, a considerable proportion of patients do not achieve adequate seizure control through pharmacological treatment~\cite{chen2018drugresistant}. Consequently, reliable seizure forecasting systems capable of providing timely warnings before seizure onset could have significant clinical and quality-of-life benefits.

Electroencephalography (EEG) has become a dominant modality for computational analysis in seizure forecasting because it provides a non-invasive measure of electrical brain activity from the scalp~\cite{rasheed2020machine, mdpi_taxonomy_2023}. Multichannel EEG recordings capture temporal and spatial patterns associated with both seizure and non-seizure brain states. Seizure evolution is typically characterized by distinct physiological phases, including interictal, preictal, ictal, and postictal periods~\cite{mdpi_taxonomy_2023, Zhang2499}. The presence of a preictal phase, which may occur minutes before seizure onset, suggests that sufficiently sensitive computational models may be able to identify transitional neural dynamics and generate clinically actionable warnings~\cite{costa2024forecasting}. However, reliable detection of preictal activity remains challenging because preictal signatures are often subtle, patient-specific, and temporally variable~\cite{frontiers_definitions_2024}.

Machine learning has emerged as a promising approach for addressing this challenge. Early studies demonstrated that classical learning algorithms can identify predictive EEG patterns and provide meaningful seizure forecasting performance~\cite{lekshmy2022comparative, diagnostics_comparison_2022, sciencedirect_comparison1}. More recent work has shown that deep learning architectures can further improve performance by learning complex temporal and spatial representations directly from EEG data~\cite{9740802, svmd_transformer_2022, zhang2024cnn_gru_am}. Temporal architectures are particularly relevant because seizure development is inherently dynamic~\cite{9740802}. Nevertheless, no single model architecture has consistently demonstrated universal superiority across patients and evaluation settings. This has motivated increasing interest in ensemble learning approaches that combine complementary model strengths~\cite{costa2024forecasting, hybrid_wide_proquest, sciencedirect_comparison2}.

Despite these advances, several methodological limitations remain in existing seizure forecasting studies. Some studies evaluate models using patients or recordings that have already contributed to training, thereby limiting conclusions about patient-independent generalization. Others insufficiently address the severe class imbalance between interictal and preictal windows, or introduce data leakage through improper splitting, calibration, or threshold-selection procedures. In addition, many pipelines do not fully incorporate clinically relevant post-processing constraints such as temporal alarm smoothing and refractory periods that are necessary for practical deployment~\cite{frontiers_definitions_2024, mdpi_taxonomy_2023}. These limitations can lead to overly optimistic estimates of forecasting performance and reduced clinical interpretability.

This study addresses these gaps through a patient-independent hybrid ensemble framework for EEG-based epileptic seizure forecasting. The main contributions of this work are as follows:
\begin{itemize}

\item We propose a hybrid stacking ensemble that integrates calibrated predictions from five deep learning and three classical machine learning models using a logistic regression meta-learner.

\item We employ a strict Leave-One-Patient-Out (LOPO) protocol to evaluate patient-independent generalization, ensuring that the held-out test patient is excluded from model training, probability calibration, meta-learning, threshold selection, and post-processing parameter selection.

\item We quantify operating-point transfer failure by comparing thresholds selected exclusively on held-out training patients with a test-tuned oracle constrained to the target false-alarm budget, and assess statistical significance using an analytic random predictor matched to each patient's realized alarm rate.


\end{itemize}

\section{Related Work}
\label{sec:litreview}

Seizure prediction and forecasting have been studied extensively using EEG, with forecasting emphasizing future seizure risk over a clinically meaningful horizon~\cite{costa2024forecasting}. CHB-MIT is one of the most widely used public benchmarks~\cite{handa2023datasets}, alongside the University of Bonn~\cite{lekshmy2022comparative, diagnostics_comparison_2022}, UCI EEG~\cite{9740802}, and Kaggle-based datasets~\cite{svmd_transformer_2022, sciencedirect_comparison2}. Prior work has explored signal filtering, normalization, dimensionality reduction, and multi-domain feature extraction~\cite{mdpi_taxonomy_2023, ieee_basic_models, hybrid_wide_proquest}, together with classical models such as support vector machines and tree-based classifiers~\cite{lekshmy2022comparative, diagnostics_comparison_2022, sciencedirect_comparison1}, and more recent convolutional, recurrent, transformer, and ensemble architectures~\cite{zhang2024cnn_gru_am, svmd_transformer_2022, electronics_cnn_fusion_2022, 9740802, costa2024forecasting, sciencedirect_comparison2}. Despite these advances, performance remains affected by limited data, class imbalance, inter-patient variability, and differences in evaluation protocols~\cite{handa2023datasets, training_size_effect, frontiers_definitions_2024}.

CHB-MIT studies vary substantially in experimental design. Truong et al.~\cite{truong2018} use CNNs with short-time Fourier representations, Gao et al.~\cite{gao2022pediatric} employ a multi-scale dilated network, Li et al.~\cite{li2023stmlp} propose a spatio-temporal MLP, and Xu et al.~\cite{xu2023drsngru} combine residual shrinkage with a GRU. However, reported results are not directly comparable because studies differ in patient partitioning, preictal horizon, alarm definition, and decision-threshold selection. Andrade et al.~\cite{andrade2024databases} further show that performance can degrade substantially when evaluation shifts from sample-level to alarm-level settings or across datasets. Many high-performing studies~\cite{truong2018, benmessaoud2021, gao2022pediatric, xu2023drsngru} are patient-specific and therefore do not directly assess generalization to unseen patients. For this reason, Table~\ref{tab:compare} separates prior work by evaluation protocol. Among patient-independent studies, Meng et al.~\cite{meng2025seresnet} report 84.4\% sensitivity at 0.232~FA/hr, providing an important reference point. Our study addresses a complementary question by examining whether an operating point selected without access to the held-out patient preserves its intended false-alarm behavior after transfer.

Evaluation methodology is therefore central to seizure forecasting. Patient-level leakage can inflate estimates of cross-patient generalization when data from the evaluated patient influence training or model selection~\cite{frontiers_definitions_2024, mdpi_taxonomy_2023}, while threshold or post-processing selection using test labels produces an oracle-like operating point unavailable in deployment. Class imbalance further complicates evaluation because preictal windows constitute only a small fraction of long-term EEG recordings~\cite{mdpi_taxonomy_2023, sciencedirect_comparison2, frontiers_definitions_2024}. Motivated by these limitations, we adopt strict LOPO evaluation, exclude the held-out patient from training and operating-point selection, report seizure-level sensitivity jointly with realized FA/hr, and compare the resulting configuration with a test-tuned oracle and a random-alarm baseline.

\section{Method}

\subsection{System Design and Data Flow}

\begin{figure}[t]
\centering
\includegraphics[width=\columnwidth, height=0.4\textheight, keepaspectratio]{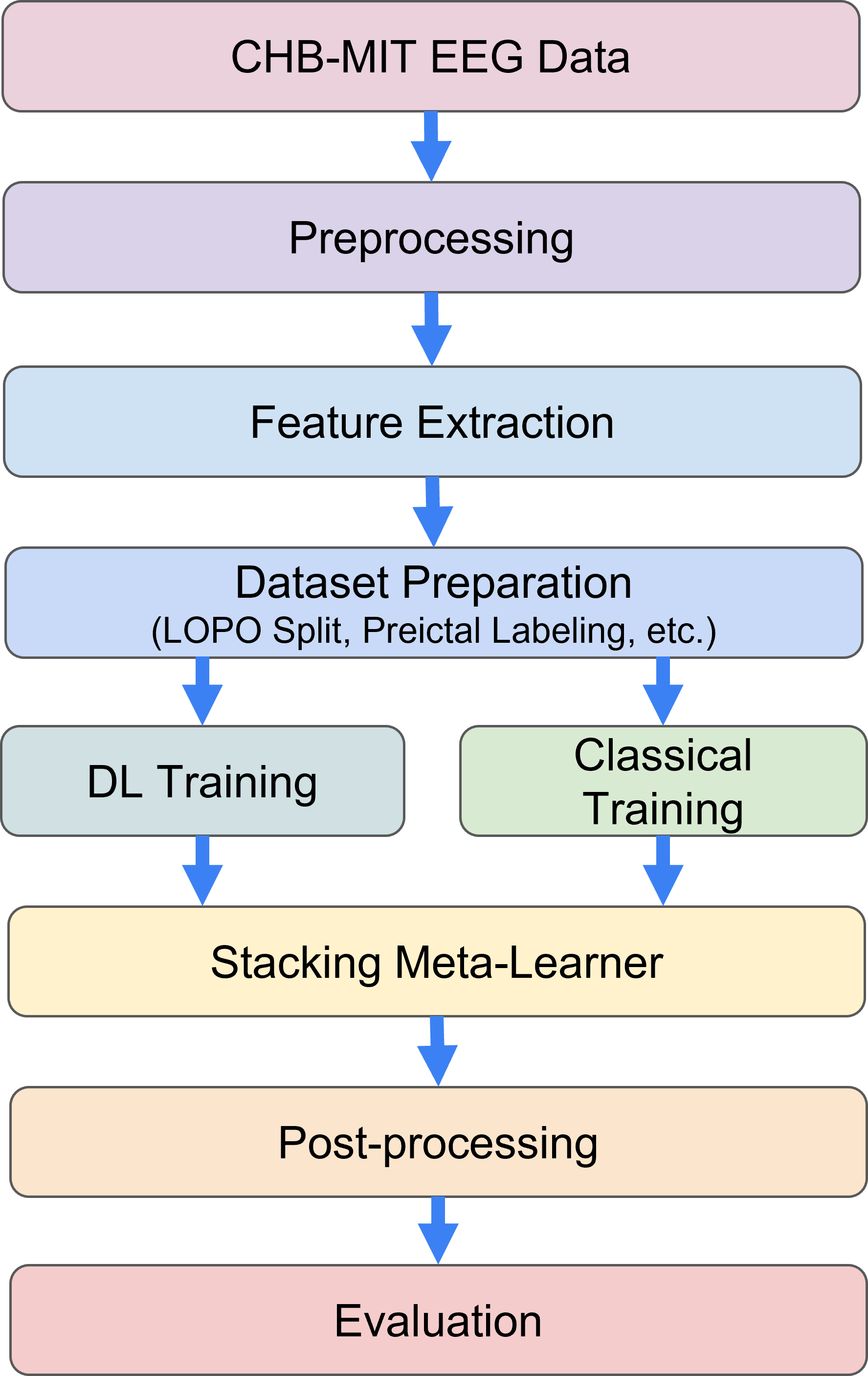}
\caption{Overview of the proposed seizure-forecasting framework, including data acquisition, preprocessing, feature extraction, LOPO-based partitioning, deep learning and classical model training, stacking-based meta-learning, temporal post-processing, and performance evaluation. Signal preprocessing is applied only to the classical machine learning branch.}
\label{fig:arch}
\end{figure}

The proposed framework processes multichannel EEG recordings using 8-second windows with a 2-second stride and produces a calibrated probability of the preictal state for each window. As illustrated in Fig.~\ref{fig:arch}, the pipeline follows a Leave-One-Patient-Out (LOPO) evaluation protocol and comprises two parallel modeling branches. The deep learning branch operates directly on multichannel EEG windows, whereas the classical machine learning branch uses PCA-compressed handcrafted features. The calibrated outputs of the constituent models are subsequently combined using a logistic regression stacking meta-learner, followed by thresholding and temporal post-processing to generate seizure alarms. For each LOPO fold, the decision threshold is selected exclusively from the held-out meta slice by evaluating 300 candidate values in the interval $[0.05,0.99]$. Across the 24 folds, the selected thresholds range from 0.58 to 0.87.

Raw \texttt{.edf} recordings are associated with the anonymized CHB-MIT patient identifiers (\texttt{chb01}--\texttt{chb24}) to support patient-independent LOPO evaluation. Recordings that cannot be successfully loaded are excluded from subsequent analysis. Preictal labeling is performed independently for each patient. For each annotated seizure, the 5-minute interval immediately preceding seizure onset is excluded to reduce ambiguity near the ictal transition, while the preceding 30-minute interval is labeled as preictal. All remaining eligible windows are labeled as interictal. This patient-specific labeling procedure preserves patient boundaries and reduces the risk of information leakage across subjects. The 5-minute exclusion interval and the preceding 30-minute preictal interval correspond to the seizure prediction horizon (SPH) and seizure occurrence period (SOP), respectively, following the terminology introduced by Maiwald et al.~\cite{maiwald2004comparison}. The SPH establishes a minimum interval between an alarm and seizure onset, thereby supporting clinically actionable forecasting while excluding EEG activity immediately adjacent to the ictal transition. The 30-minute SOP follows commonly adopted CHB-MIT forecasting configurations~\cite{truong2018,gao2022pediatric} and is fixed a priori rather than optimized on the evaluation data.



\subsection{EEG Preprocessing}

Signal preprocessing is applied exclusively to the handcrafted-feature branch, while the deep learning branch operates directly on resampled EEG windows. Prior to feature extraction, each EEG channel undergoes linear detrending to remove baseline drift, followed by a fourth-order Butterworth bandpass filter spanning 0.5--45\,Hz. Zero-phase filtering is implemented using \texttt{filtfilt} to minimize phase distortion. A 60\,Hz IIR notch filter is subsequently applied for consistency with the implemented preprocessing pipeline, although the preceding 45\,Hz low-pass cutoff substantially attenuates power-line interference at 60\,Hz. Finally, a median filter with a kernel size of 5 is used to suppress residual impulsive artifacts.



\subsection{Feature Extraction}

Handcrafted features are extracted from each channel within every 8-second EEG window. Frequency-domain features comprise band power estimated using Welch's method in the delta (0.5--4\,Hz), theta (4--8\,Hz), alpha (8--13\,Hz), beta (13--30\,Hz), and low-gamma (30--45\,Hz) bands. Additional features include coefficient statistics derived from a four-level Daubechies-4 wavelet decomposition, line length, spectral entropy, Hjorth parameters, and cross-channel correlation statistics characterizing spatial synchrony. Using the set of common channels available for each recording yields 30 features per channel together with six cross-channel correlation summaries. Consequently, feature-vector dimensionality ranges from 126 to 906 across recordings; vectors with fewer dimensions are zero-padded to 906 dimensions prior to PCA.

Because the preprocessing pipeline limits the analyzed frequency range to 45\,Hz, the extracted spectral features characterize only the low-gamma band and do not capture higher-frequency gamma activity. This frequency-range restriction is considered a limitation of the present study. Feature extraction is parallelized using \texttt{joblib}, and the resulting feature vectors are cached to disk to avoid redundant computation across subsequent LOPO folds.

\subsection{Leave-One-Patient-Out Cross-Validation}

We employ a strict Leave-One-Patient-Out (LOPO) protocol in which one patient is held out exclusively for testing in each fold, while the remaining 23 patients form the development pool. Using a fixed random seed of 42, the development data are partitioned into non-overlapping subsets comprising 70\% for base-model training, 7.5\% for validation, 7.5\% for probability calibration, and 15\% for training the logistic regression meta-learner. Boundary windows are discarded to prevent overlap between adjacent subsets. The held-out patient is therefore excluded from model training, calibration, threshold selection, and post-processing parameter selection.

\subsection{Base Models}

The deep learning branch is trained with AdamW, OneCycleLR, automatic mixed precision, gradient clipping, and binary cross-entropy with logits using \texttt{pos\_weight}=3.0. Class imbalance is handled through 3:1 interictal undersampling, capped at 400 preictal windows per patient, and weighted loss. Deep models are trained only when at least 30 preictal windows are available; otherwise, the fold uses only classical models. Positive samples are augmented with Gaussian noise ($\sigma=0.05$) with 50\% probability.

We evaluate five deep learning architectures: EEGNet, BiLSTM, Temporal Fusion Transformer (TFT), Conformer, and Dilated TCN. The Dilated TCN uses six residual blocks with dilation rates from \(2^0\) to \(2^5\), yielding a receptive field of 253 samples. The classical branch uses PCA-compressed features with Random Forest, XGBoost, and LinearSVC. Random Forest uses 100 trees with max depth 15 and balanced weights. XGBoost uses 500 trees, learning rate 0.05, maximum depth 7, and \texttt{scale\_pos\_weight} based on the interictal-to-preictal ratio. LinearSVC is wrapped in \texttt{CalibratedClassifierCV} to produce calibrated predictions.

\subsection{Calibration, Stacking, and Post-Processing}
\label{sec:postproc}

All base-model outputs are calibrated before stacking. Deep learning models use Platt scaling on the calibration slice, Random Forest and XGBoost use \texttt{CalibratedClassifierCV}, and LinearSVC is calibrated during three-fold fitting. Calibrated probabilities from active models then stack and train a logistic regression meta-learner on the meta slice. Deep learning models with AUC below 0.55 are excluded, and meta-learner coefficients are recorded to assess model contributions.

The meta-learner outputs a continuous preictal probability for each window, which is converted to binary predictions using a decision threshold. Temporal smoothing confirms an alarm only after a required number of positives within a sliding window, while refractory enforcement imposes a silence period after each alarm to reduce repeated warnings.

All thresholds and post-processing parameters are selected on the held-out meta set rather than test labels. Candidate settings include 3-of-5, 4-of-7, and 5-of-10 smoothing with 30-, 45-, and 60-minute refractory periods. The final configuration uses 4-of-7 smoothing with a 30-minute refractory period, achieving 0.93 sensitivity and 0.95 FA/hr on the meta set. All test results use this fixed configuration.

\section{Experiments}

\textbf{Dataset.} We evaluate the proposed framework on the CHB-MIT scalp-EEG dataset~\cite{goldberger2000physiobank}, a widely used benchmark for seizure forecasting. The dataset contains 22 subjects organized into 24 patient collections, which are treated as independent evaluation units in this study. It comprises approximately 42~GB of data across 664 \texttt{.edf} files sampled at 256~Hz with 16-bit resolution and recorded according to the International 10--20 electrode placement system. Most recordings contain 23 EEG channels, with a small number containing 24 or 26 channels; some also include ECG or VNS signals. Seizure annotations are provided for 129 files, corresponding to 198 annotated seizures. Across 1{,}766{,}524 extracted windows, 8.30\% are labeled as preictal, highlighting the substantial class imbalance in the dataset.

\textbf{Configuration.} Experiments are implemented in Python~3 using a Google Colab L4 GPU runtime. PyTorch with automatic mixed precision is used for deep learning models, XGBoost is trained in GPU histogram mode, and scikit-learn is used for classical models, calibration, PCA, and evaluation metrics. Signal processing is performed using SciPy and PyWavelets. All random seeds are fixed at 42 for Python, NumPy, and PyTorch, with \texttt{cudnn.deterministic} enabled. 

\textbf{Protocol.} All experiments follow a strict Leave-One-Patient-Out (LOPO) cross-validation protocol across the 24 CHB-MIT patient collections~\cite{goldberger2000physiobank}. A 30-minute preictal horizon and a 5-minute exclusion gap before seizure onset are used, consistent with clinically motivated forecasting windows~\cite{frontiers_definitions_2024}. Performance is evaluated at both the window level and seizure level. Window-level evaluation uses individual 8-second windows with a 2-second stride, whereas seizure-level evaluation measures whether at least one alarm occurs within the corresponding preictal block. The primary metrics are seizure-level sensitivity, false alarms per hour (FA/hr), and warning time. Warning time is defined as the duration from the first detection to the end of the labeled preictal block; the actual lead time to seizure onset is larger by the fixed 5-minute exclusion gap. Window-level sensitivity, specificity, precision, F1-score, and AUC-ROC are reported for additional context.

\subsection{Seizure-Level Detection Performance}

Table~\ref{tab:perpatient} summarizes seizure-level performance under LOPO cross-validation. Across all 24 patients, the proposed framework detects 97 of 141 seizures, achieving 72.8\% sensitivity, a mean false-alarm rate of 1.251~FA/hr, and a mean warning time of 18.3 minutes. Four patients (\texttt{chb12}, \texttt{chb15}, \texttt{chb16}, and \texttt{chb24}) exhibit anomalous preictal proportions below 1\% or above 15\% and are excluded from the filtered aggregate. On the resulting cohort of 20 patients and 100 seizures, the framework detects 69 seizures, yielding a mean sensitivity of $0.742 \pm 0.256$, a mean false-alarm rate of $1.244 \pm 0.451$~FA/hr, and a mean warning time of $16.9 \pm 7.0$ minutes.

The realized false-alarm rate exceeds the 1.0~FA/hr target selected on the meta set, indicating imperfect transfer of the operating point to unseen patients. Specifically, a threshold satisfying the target false-alarm constraint on held-out development data does not necessarily preserve the same operating characteristics for a previously unseen patient.

To assess whether the observed seizure-level performance exceeds chance, we compare the framework with an unspecific random predictor following established seizure-prediction evaluation practice~\cite{winterhalder2003spc, schelter2006significance}. Under a Poisson alarm process matched to each patient's realized false-alarm rate, the probability of detecting a seizure within a prediction interval of duration $T$ is $1-\exp(-\lambda T)$, where $\lambda$ denotes FA/hr and $T=0.5$~h. The corresponding expected sensitivities are 44.6\% for the full cohort and 44.9\% for the filtered cohort, compared with observed sensitivities of 72.8\% and 74.2\%, respectively. For all 141 seizures, the resulting Poisson-binomial null predicts $60.4 \pm 5.5$ detections, whereas 97 are observed ($z=6.55$, one-sided exact $p=1.6\times10^{-11}$). Similarly, the filtered cohort yields 69 observed detections compared with 42.6 expected under the null ($p=3.1\times10^{-8}$), demonstrating performance significantly above chance.

\begin{table}[t]
\centering
\caption{Per-patient seizure-level performance under LOPO cross-validation using a 30-minute preictal horizon and 5-minute exclusion gap. \textdagger{} denotes patients with anomalous preictal rates excluded from the filtered aggregate. 
}
\label{tab:perpatient}
\begin{tabular}{@{}lrcrr@{}}
\toprule
Patient & Sens. & Detected & FA/hr & Warn (min) \\
\midrule
chb01 & 28.6\% & 2/7 & 1.16 & 15.5 \\
chb02 & 100.0\% & 2/2 & 1.24 & 10.3 \\
chb03 & 28.6\% & 2/7 & 0.52 & 9.6 \\
chb04 & 100.0\% & 4/4 & 1.46 & 12.7 \\
chb05 & 60.0\% & 3/5 & 0.74 & 18.3 \\
chb06 & 50.0\% & 5/10 & 0.57 & 13.7 \\
chb07 & 66.7\% & 2/3 & 0.56 & 9.3 \\
chb08 & 100.0\% & 5/5 & 1.43 & 17.7 \\
chb09 & 75.0\% & 3/4 & 0.96 & 24.3 \\
chb10 & 57.1\% & 4/7 & 0.77 & 5.2 \\
chb11 & 100.0\% & 3/3 & 1.78 & 18.8 \\
chb12$^\dagger$ & 0.0\% & 0/10 & 0.00 & N/A \\
chb13 & 87.5\% & 7/8 & 1.28 & 19.8 \\
chb14 & 66.7\% & 4/6 & 1.17 & 23.0 \\
chb15$^\dagger$ & 84.6\% & 11/13 & 1.56 & 23.3 \\
chb16$^\dagger$ & 80.0\% & 4/5 & 1.73 & 45.3 \\
chb17 & 100.0\% & 3/3 & 1.85 & 7.4 \\
chb18 & 100.0\% & 5/5 & 2.01 & 16.7 \\
chb19 & 33.3\% & 1/3 & 1.30 & 25.2 \\
chb20 & 50.0\% & 3/6 & 1.34 & 27.3 \\
chb21 & 100.0\% & 4/4 & 1.79 & 11.2 \\
chb22 & 100.0\% & 3/3 & 1.87 & 18.8 \\
chb23 & 80.0\% & 4/5 & 1.07 & 32.7 \\
chb24$^\dagger$ & 100.0\% & 13/13 & 1.86 & 14.7 \\
\midrule
Mean (all, $n{=}24$) & 72.8\% & 97/141 & 1.251 & 18.3 \\
Mean (filt., $n{=}20$) & 74.2\% & 69/100 & 1.244 & 16.9 \\
\bottomrule
\end{tabular}
\end{table}

\subsection{Post-Processing Operating Characteristics}

\begin{figure}[t]
\centering
\includegraphics[width=\columnwidth]{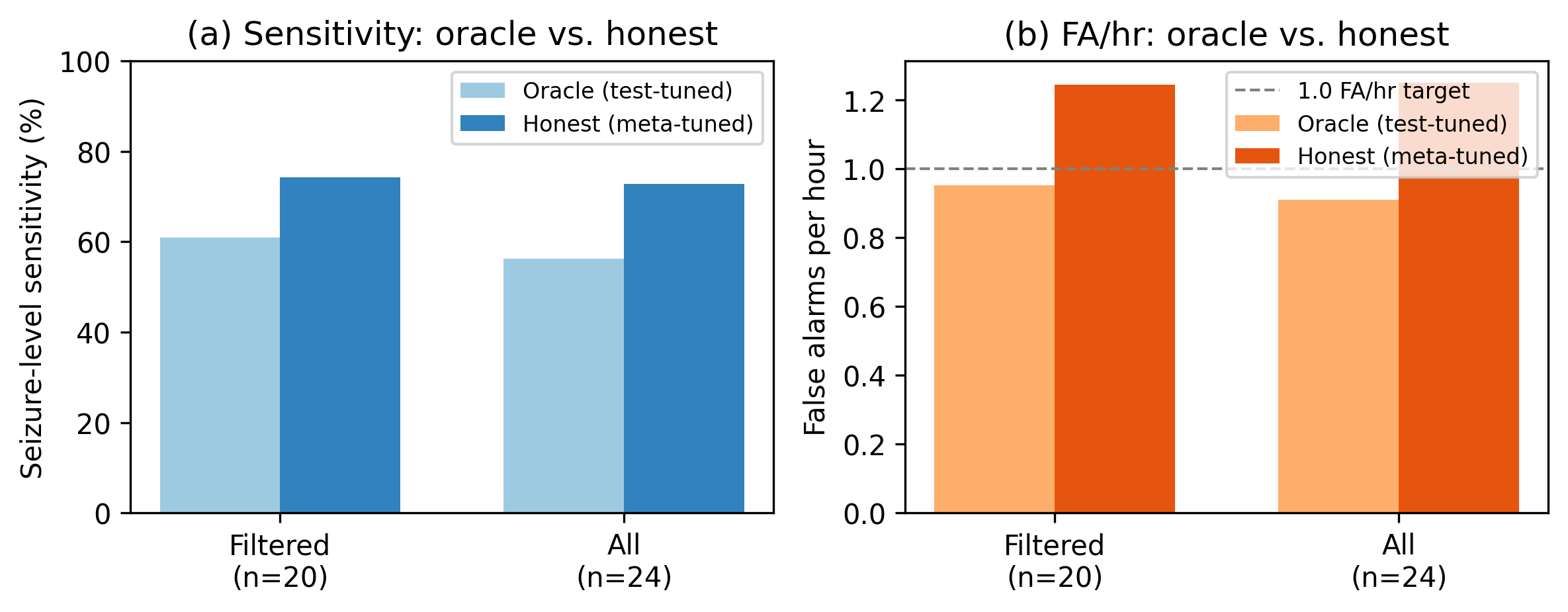}
\caption{Sensitivity and FA/hr for meta-tuned and test-tuned configurations under fixed post-processing settings.}
\label{fig:curves}
\end{figure}

Fig.~\ref{fig:curves} compares a leakage-controlled (control) operating point with an oracle configuration that tunes the decision threshold with each test patient's labels. Both settings use the same 4-of-7 smoothing rule and 30-minute refractory period. The control configuration, whose threshold is selected on the meta set, achieves 74.2\% sensitivity on the filtered cohort at 1.244~FA/hr, exceeding the 1.0~FA/hr target. In contrast, the oracle configuration, constrained to $\leq$1.0~FA/hr and allowed access to test labels, achieves 60.9\% sensitivity at 0.951~FA/hr.

This 13.3 percentage-point difference reflects operating-point transfer failure: thresholds selected from held-out training patients are too permissive when applied to unseen patients, producing both more detections and more false alarms. Therefore, reporting sensitivity without its realized false-alarm rate would overstate clinical utility. The oracle acts as a reference for the performance achievable under the stated false-alarm constraint, but it is not usable because it trains on test labels. 

\subsection{Window-Level Classification}

Although seizure-level metrics are the primary basis for evaluation, window-level metrics provide additional insight into classifier behavior. As shown in Table~\ref{tab:window}, the decision threshold is selected per fold to maximize seizure-level sensitivity subject to a 1.0~FA/hr ceiling on the meta set, with F1-score optimization used only when the target cannot be satisfied. Mean specificity remains near 0.999, indicating that individual interictal windows are rarely flagged. However, window-level AUC-ROC remains near chance, with means of $0.487 \pm 0.085$ across all patients and $0.485 \pm 0.091$ on the filtered cohort.

This near-chance AUC should be interpreted carefully. AUC evaluates rank ordering across individual windows, whereas the forecasting objective is event-level alarm generation over extended preictal intervals. Only chb08 exceeds 0.600 window-level AUC, reaching 0.726 and achieving 100\% seizure-level sensitivity. Overall, no strong general relationship is observed between window-level AUC and seizure-level coverage. The low window-level sensitivity, precision, and F1-score are expected consequences of the thresholding and post-processing strategy, which intentionally suppresses frequent window-level positives to reduce false alarms.

\begin{table}[t]
\centering
\caption{Average window-level performance, reported as mean $\pm$ standard deviation for all and filtered patients.}
\label{tab:window}

\begin{tabular}{l l l }
\toprule
Metric & All ($n{=}24$) & Filtered ($n{=}20$) \\
\midrule
Sensitivity  & $0.001 \pm 0.000$ & $0.001 \pm 0.000$ \\
Specificity  & $0.999 \pm 0.000$ & $0.999 \pm 0.000$ \\
Precision    & $0.104 \pm 0.077$ & $0.086 \pm 0.046$ \\
F1-score     & $0.002 \pm 0.001$ & $0.002 \pm 0.001$ \\
Accuracy     & $0.890 \pm 0.092$ & $0.925 \pm 0.040$ \\
AUC-ROC      & $0.487 \pm 0.085$ & $0.485 \pm 0.091$ \\
\bottomrule
\end{tabular}
\end{table}

\subsection{Protocol-Aware Comparison}

Table~\ref{tab:compare} compares the proposed pipeline with prior CHB-MIT results under patient-specific and patient-independent protocols. Patient-independent LOPO results are the most relevant because the test patient is fully excluded from training. Under this setting, Meng et al.~\cite{meng2025seresnet} report 84.4\% sensitivity at 0.232~FA/hr, while our leakage-controlled model achieves 74.2\% sensitivity at 1.244~FA/hr on the filtered cohort. Patient-specific methods are included only for context, as they train and test within the subject and thus address an easier task.

\begin{table}[!ht]
\centering
\caption{Comparison of CHB-MIT seizure forecasting results across evaluation protocols \protect\footnotemark.}
\label{tab:compare}
\begin{tabular}{lcccc}
\toprule
Method & Year & Sens. & FA/hr & Threshold \\
\midrule
\multicolumn{5}{@{}l}{\emph{Patient-specific evaluation}}\\
Truong et al.~\cite{truong2018} & 2018 & 0.812 & 0.159 & N/R \\
Ben Messaoud \& Chavez~\cite{benmessaoud2021} & 2021 & 0.821 & 0.080 & N/R \\
\midrule
\multicolumn{5}{@{}l}{\emph{Patient-independent LOPO evaluation}}\\
Meng et al.\ (3D-SERESNet)~\cite{meng2025seresnet} & 2025 & 0.844 & 0.232 & N/R \\
Ours, honest (all, $n{=}24$) & 2026 & 0.728 & 1.251 & Meta slice \\
Ours, honest (filt., $n{=}20$) & 2026 & 0.742 & 1.244 & Meta slice \\
Ours, oracle (filt., $n{=}20$) & 2026 & 0.609 & 0.951 & Test labels \\
\bottomrule
\end{tabular}
\end{table}

\footnotetext{The threshold-source column indicates whether the reported operating point was selected using a held-out meta slice, test labels, or was not reported. Results across protocol groups are not directly comparable.}

\section{Analysis and Discussion}
\label{sec:analysis}

The model-level analysis indicates substantial variation in cross-patient forecasting performance across architectures. Among the deep learners, the Conformer most consistently exceeded an AUC of 0.55, whereas TFT and Dilated TCN contributed intermittently and EEGNet and BiLSTM remained near chance. The Conformer also received consistently positive meta-learner weights, suggesting that its combination of convolutional and attention-based representations was better suited to capturing complementary local and long-range EEG dynamics~\cite{zhang2024cnn_gru_am}. The classical models also contributed meaningfully, while some tree-based predictions exhibited partial redundancy. Collectively, these results indicate that the Conformer and classical branch accounted for most of the useful ensemble signal, and therefore do not provide strong empirical support for retaining all deep architectures in the final ensemble.

Because exhaustive leave-one-model-out ablation would require retraining the deep-learning branch for multiple model subsets, we instead analyze per-fold model admission and meta-learner weights recorded during training. Although this does not replace a formal ablation study, it provides an indication of the relative contribution of each base learner. Performance also varies considerably across patients: \texttt{chb01} and \texttt{chb03} achieve only 28.6\% seizure-level sensitivity, whereas 8 of the 20 patients in the filtered cohort achieve 100\%. This heterogeneity suggests that a single global ensemble may not generalize uniformly across patients and motivates future investigation of patient-adaptive model selection and calibration.

Window-level metrics should be interpreted in the context of the event-level alarm-generation objective. The low window-level sensitivity, precision, and F1-score arise from conservative thresholding, 4-of-7 temporal smoothing, and the 30-minute refractory period, which suppress isolated positive predictions and produce specificity near 0.999. Similarly, the near-chance AUC-ROC reflects the difficulty of discriminating individual preictal and interictal windows under patient-independent evaluation. In contrast, seizure forecasting is evaluated primarily at the event level, where performance depends on whether temporally aggregated model outputs generate an alarm within the predefined preictal interval. These results therefore reinforce the importance of interpreting window-level discrimination jointly with seizure-level sensitivity and false-alarm burden.

\section{Limitations and Future Work}

This study is limited by the size of the CHB-MIT dataset, the computational cost of the proposed pipeline, and the inherent difficulty of patient-independent seizure forecasting. A complete experimental run required approximately 32 hours on an NVIDIA L4 GPU, which constrained exhaustive hyperparameter tuning, external dataset validation, and broader optimization of AUC-ROC and false-alarm performance.

A further limitation arises from the internal data partitioning strategy. The deep-learning training pool is indexed using the unshuffled patient order, whereas the calibration and meta-learning subsets are derived from the shuffled ordering. Consequently, the training pool overlaps with 71\% of the calibration slice and 93\% of the meta slice at the patient level. Although balanced sampling limits each patient to at most 400 preictal windows, allowing no more than approximately 14\% of the meta slice to enter deep-model training, this overlap may produce optimistic deep-model probabilities on the meta set. Importantly, the held-out LOPO test patient remains fully excluded from all model-development stages; thus, the reported test results remain patient-independent. Future work should adopt strictly disjoint patient-level partitions for training, validation, calibration, and meta-learning.

Future research will focus on improving false-alarm control, operating-point transfer, and patient adaptation. Patient-adaptive ensemble weighting, threshold calibration, and post-processing may improve performance when limited patient-specific calibration data are available. Evaluation on larger and more diverse EEG datasets will also be necessary to assess robustness, generalizability, and clinical applicability.

\section{Conclusion}

This study presents a hybrid ensemble framework for patient-independent epileptic seizure forecasting using CHB-MIT EEG data. Under LOPO evaluation, the filtered cohort achieves 74.2\% seizure-level sensitivity at 1.24~FA/hr, detecting 69 of 100 seizures with an average warning time of 16.9 minutes. The results also reveal an operating-point transfer limitation: although the meta-selected configuration targets 1.0~FA/hr, the realized false-alarm rate increases to 1.24~FA/hr on unseen patients, while a test-tuned oracle constrained to the target budget achieves 60.9\% sensitivity at 0.951~FA/hr. These findings highlight the importance of reporting seizure-level sensitivity together with the realized false-alarm rate and clearly documenting the procedure used for threshold selection in patient-independent seizure forecasting.


\section*{Acknowledgments} \label{Acknowledgements}
The authors acknowledge the use of AI tools for grammar checking, proofreading, and code debugging assistance. All technical content, analyses, and conclusions were reviewed and verified by the authors.


\bibliographystyle{IEEEtran}
\bibliography{references}

@article{handa2023datasets,
  author  = {Handa, Palak and Mathur, Monika and Goel, Nidhi},
  title   = {EEG Datasets in Machine Learning Applications of Epilepsy Diagnosis and Seizure Detection},
  journal = {SN Computer Science},
  volume  = {4},
  pages   = {437},
  year    = {2023},
  doi     = {10.1007/s42979-023-01958-z},
  url     = {https://link.springer.com/article/10.1007/s42979-023-01958-z}
}

@article{costa2024forecasting,
  author  = {Costa, Gon{\c{c}}alo and Teixeira, C{\'e}sar Alexandre and Pinto, Mauro},
  title   = {Comparison between epileptic seizure prediction and forecasting based on machine learning},
  journal = {Scientific Reports},
  volume  = {14},
  year    = {2024},
  doi     = {10.1038/s41598-024-56019-z},
  url     = {https://www.nature.com/articles/s41598-024-56019-z}
}

@inproceedings{lekshmy2022comparative,
  author    = {Lekshmy, H. O. and Panickar, Dhanyalaxmi and Harikumar, Sandhya},
  title     = {Comparative Analysis of Multiple Machine Learning Algorithms for Epileptic Seizure Prediction},
  booktitle = {Journal of Physics: Conference Series},
  volume    = {2161},
  number    = {1},
  pages     = {012055},
  year      = {2022},
  doi       = {10.1088/1742-6596/2161/1/012055}
}

@phdthesis{hybrid_wide_proquest,
  author = {Altaf, Zarqa and Unar, Mukhtiar Ali and Narejo, Sanam and Zaki, Muhammad Ahmed and Naseer-u-Din},
  title  = {Generalized Epileptic Seizure Prediction Using Machine Learning Method},
  school = {ProQuest Dissertations and Theses},
  year   = {2023},
  url    = {https://www.proquest.com/openview/e3a21d7a8257da65420fa0c0aeba8889}
}

@inproceedings{ieee_basic_models,
  author    = {Talukder, Md. Simul Hasan and Sulaiman, Rejwan Bin},
  title     = {Comparative Analysis of Epileptic Seizure Prediction: Exploring Diverse Pre-processing Techniques and Machine Learning Models},
  booktitle = {2023 10th IEEE Uttar Pradesh Section International Conference on Electrical, Electronics and Computer Engineering},
  year      = {2023},
  doi       = {10.1109/10434289}
}

@article{svmd_transformer_2022,
  author  = {Wu, Xiao and Zhang, Tinglin and Zhang, Limei and Qiao, Lishan},
  title   = {Epileptic Seizure Prediction Using Successive Variational Mode Decomposition and Transformers Deep Learning Network},
  journal = {Frontiers in Neuroscience},
  volume  = {16},
  pages   = {982541},
  year    = {2022},
  doi     = {10.3389/fnins.2022.982541}
}

@article{zhang2024cnn_gru_am,
  author  = {Zhang, Jincan and Zheng, Shaojie and Chen, Wenna and Du, Ganqin and Fu, Qizhi and Jiang, Hongwei},
  title   = {A Scheme Combining Feature Fusion and Hybrid Deep Learning Models for Epileptic Seizure Detection and Prediction},
  journal = {Scientific Reports},
  volume  = {14},
  pages   = {67855},
  year    = {2024},
  doi     = {10.1038/s41598-024-67855-4}
}

@article{electronics_cnn_fusion_2022,
  author  = {Ouichka, O. and others},
  title   = {Deep Learning Models for Predicting Epileptic Seizures Using iEEG Signals},
  journal = {Electronics},
  volume  = {11},
  number  = {4},
  pages   = {605},
  year    = {2022},
  doi     = {10.3390/electronics11040605}
}

@inproceedings{9740802,
  author    = {Nanthini, K. and Tamilarasi, A. and Pyingkodi, M. and Dishanthi, M. and Kaviya, S. M. and Mohideen, P. Aslam},
  title     = {Epileptic Seizure Detection and Prediction Using Deep Learning Technique},
  booktitle = {2022 International Conference on Computer Communication and Informatics (ICCCI)},
  pages     = {1--7},
  year      = {2022},
  doi       = {10.1109/ICCCI54379.2022.9740802}
}

@article{sciencedirect_comparison1,
  author  = {Almustafa, Khaled Mohamad},
  title   = {Classification of Epileptic Seizure Dataset Using Different Machine Learning Algorithms},
  journal = {Informatics in Medicine Unlocked},
  year    = {2020},
  url     = {https://www.sciencedirect.com/science/article/pii/S2352914820305943}
}

@article{sciencedirect_comparison2,
  author  = {Usman, Syed Muhammad and Khalid, Shehzad and Bashir, Sadaf},
  title   = {A Deep Learning Based Ensemble Learning Method for Epileptic Seizure Prediction},
  journal = {Computers in Biology and Medicine},
  year    = {2021},
  url     = {https://www.sciencedirect.com/science/article/pii/S0010482521005047}
}

@article{diagnostics_comparison_2022,
  author  = {Tran, Ly V. and Tran, Hieu M. and Le, Tuan M. and Huynh, Tri T. M. and Tran, Hung T. and Dao, Son V. T.},
  title   = {Application of Machine Learning in Epileptic Seizure Detection},
  journal = {Diagnostics},
  volume  = {12},
  number  = {11},
  pages   = {2879},
  year    = {2022},
  doi     = {10.3390/diagnostics12112879}
}

@article{mdpi_taxonomy_2023,
  author  = {Farooq, Muhammad Shoaib and Zulfiqar, Aimen and Riaz, Shamyla},
  title   = {Epileptic Seizure Detection Using Machine Learning: Taxonomy, Opportunities, and Challenges},
  journal = {Diagnostics},
  volume  = {13},
  number  = {6},
  pages   = {1058},
  year    = {2023},
  doi     = {10.3390/diagnostics13061058}
}

@article{frontiers_definitions_2024,
  author  = {Kerr, Wesley T. and McFarlane, Katherine N. and Pucci, Gabriela Figueiredo},
  title   = {The Present and Future of Seizure Detection, Prediction, and Forecasting with Machine Learning, Including the Future Impact on Clinical Trials},
  journal = {Frontiers in Neurology},
  year    = {2024},
  doi     = {10.3389/fneur.2024.1425490}
}

@article{training_size_effect,
  author  = {Halimeh, Mustafa and Jackson, Michele and Loddenkemper, Tobias and Meisel, Christian},
  title   = {Training Size Predictably Improves Machine Learning-Based Epileptic Seizure Forecasting from Wearables},
  journal = {Neuroscience Informatics},
  year    = {2025},
  url     = {https://www.sciencedirect.com/science/article/pii/S2772528624000293}
}

@article{goldberger2000physiobank,
  author  = {A. L. Goldberger and L. A. N. Amaral and L. Glass and 
             J. M. Hausdorff and P. Ch. Ivanov and R. G. Mark and 
             J. E. Mietus and G. B. Moody and C.-K. Peng and H. E. Stanley},
  title   = {PhysioBank, PhysioToolkit, and PhysioNet: Components of a New Research Resource for Complex Physiologic Signals},
  journal = {Circulation},
  volume  = {101},
  number  = {23},
  pages   = {e215--e220},
  year    = {2000},
  doi     = {10.1161/01.CIR.101.23.e215}
}

@article {Zhang2499,
	author = {Zhang, Zhang J. and Koifman, Julius and Shin, Damian S. and Ye, Hui and Florez, Carlos M. and Zhang, Liang and Valiante, Taufik A. and Carlen, Peter L.},
	title = {Transition to Seizure: Ictal Discharge Is Preceded by Exhausted Presynaptic GABA Release in the Hippocampal CA3 Region},
	volume = {32},
	number = {7},
	pages = {2499--2512},
	year = {2012},
	doi = {10.1523/JNEUROSCI.4247-11.2012},
	publisher = {Society for Neuroscience},
	issn = {0270-6474},
	URL = {https://www.jneurosci.org/content/32/7/2499},
	eprint = {https://www.jneurosci.org/content/32/7/2499.full.pdf},
	journal = {Journal of Neuroscience}
}

@article{devinsky2016sudep,
  author  = {Devinsky, Orrin and Hesdorffer, Dale C. and Thurman,
             David J. and Lhatoo, Samden and Richerson, George},
  title   = {Sudden Unexpected Death in Epilepsy: Epidemiology,
             Mechanisms, and Prevention},
  journal = {Lancet Neurology},
  volume  = {15},
  number  = {10},
  pages   = {1075--1088},
  year    = {2016},
  note    = {PubMed PMID: 27571159},
  url     = {https://pubmed.ncbi.nlm.nih.gov/27571159/}
}

@article{truong2018,
  author  = {Truong, Nhan Duy and others},
  title = {Convolutional neural networks for seizure prediction using intracranial and scalp electroencephalogram},
  journal = {Neural Networks},
  volume = {105},
  pages = {104-111},
  year = {2018},
  issn = {0893-6080},
  doi = {https://doi.org/10.1016/j.neunet.2018.04.018},
  url = {https://www.sciencedirect.com/science/article/pii/S0893608018301485}
}

@article{rasheed2020machine,
  title={Machine learning for predicting epileptic seizures using EEG signals: A review},
  author={Rasheed, Khansa and Qayyum, Adnan and Qadir, Junaid and Sivathamboo, Shobi and Kwan, Patrick and Kuhlmann, Levin and O’Brien, Terence and Razi, Adeel},
  journal={IEEE reviews in biomedical engineering},
  volume={14},
  pages={139--155},
  year={2020},
  publisher={IEEE}
}

@article{benmessaoud2021,
  author        = {Ben Messaoud, R{\'e}my and Chavez, Mario},
  title         = {Random Forest Classifier for {EEG}-Based Seizure Prediction},
  journal       = {arXiv preprint},
  volume        = {arXiv:2106.04510},
  year          = {2021},
  month         = {June},
  url           = {https://arxiv.org/abs/2106.04510}
}

@article{meng2025seresnet,
  author        = {Meng, Lu and Zhou, Lijun and others},
  title         = {Robust Epileptic Seizure Prediction: A {3D-SERESNet} Framework
                   for Patient-Specific and Multi-Patient Generalization},
  journal       = {iScience},
  volume        = {28},
  number        = {12},
  pages         = {114171},
  year          = {2025},
  month         = {November},
  publisher     = {Elsevier},
  doi           = {10.1016/j.isci.2025.114171},
  note          = {PMC12723167}
}

@article{gao2022pediatric,
  author={Y. Gao and X. Chen and A. Liu and D. Liang and L. Wu and R. Qian and H. Xie and Y. Zhang}, 
  title={Pediatric Seizure Prediction in Scalp {EEG} Using a Multi-Scale Neural Network With Dilated Convolutions},
  journal={IEEE J. Transl. Eng. Health Med.},
  volume={10},
  pages={4900209},
  year={2022}
}

@article{li2023stmlp, 
  author={C. Li and others},
  title={Spatio-temporal {MLP} network for seizure prediction using {EEG} signals},
  journal={Measurement},
  volume={206},
  pages={112278},
  year={2023}
}

@article{xu2023drsngru,
  author={X. Xu and others},
  title={Patient-specific method for predicting epileptic seizures based on {DRSN-GRU}},
  journal={Biomed. Signal Process. Control},
  volume={81}, pages={104449}, year={2023}
}

@article{andrade2024databases,
  author={A. Andrade and C. A. Teixeira and M. Pinto},
  title={On the performance of seizure prediction machine learning methods across different databases: the sample and alarm-based perspectives},
  journal={Frontiers in Neuroscience},
  volume={18},
  pages={1417748},
  year={2024}
}

@article{maiwald2004comparison,
  author={T. Maiwald and M. Winterhalder and R. Aschenbrenner-Scheibe and H. U. Voss and A. Schulze-Bonhage and J. Timmer},
  title={Comparison of three nonlinear seizure prediction methods by means of the seizure prediction characteristic},
  journal={Physica D},
  volume={194},
  number={3--4},
  pages={357--368},
  year={2004}
}

@article{chen2018drugresistant,
  author={Z. Chen and M. J. Brodie and D. Liew and P. Kwan},
  title={Treatment Outcomes in Patients With Newly Diagnosed Epilepsy Treated With Established and New Antiepileptic Drugs: A 30-Year Longitudinal Cohort Study},
  journal={JAMA Neurology},
  volume={75},
  number={3},
  pages={279--286},
  year={2018}
}

@article{winterhalder2003spc, 
  author={M. Winterhalder and T. Maiwald and H. U. Voss and R. Aschenbrenner-Scheibe and J. Timmer and A. Schulze-Bonhage}, 
  title={The seizure prediction characteristic: a general framework to assess and compare seizure prediction methods}, 
  journal={Epilepsy \& Behavior},
  volume={4}, 
  number={3},
  pages={318--325},
  year={2003}
}

@article{schelter2006significance,
  author={B. Schelter and M. Winterhalder and T. Maiwald and A. Brandt and A. Schad and A. Schulze-Bonhage and J. Timmer}, 
  title={Testing statistical significance of multivariate time series analysis techniques for epileptic seizure prediction}, 
  journal={Chaos}, 
  volume={16}, 
  number={1}, 
  pages={013108}, 
  year={2006}
}

\end{document}